\documentclass[letterpaper, 10 pt, conference]{ieeeconf}  
\IEEEoverridecommandlockouts    
\usepackage[utf8]{inputenc}
\usepackage[english]{babel}
\usepackage{times} 

\usepackage{amssymb,amsmath,amsthm}
\usepackage{cite}
\usepackage{multirow}
\usepackage{graphicx} 
\usepackage[table,xcdraw]{xcolor}
\usepackage{psfrag}
\usepackage{color}
\usepackage[us]{datetime}
\usepackage{algorithm}
\usepackage{algorithmic}
\usepackage{balance}
\usepackage{makecell}
\usepackage{mathtools}
\usepackage{array}
\usepackage{lipsum}
\usepackage{float}
\usepackage{color}
\usepackage{hyperref}
\usepackage{cleveref}
\usepackage{siunitx}
\usepackage[export]{adjustbox}
\usepackage{placeins}

\usepackage{subcaption}

\DeclareCaptionLabelSeparator{periodspace}{.\quad}
\renewcommand{\baselinestretch}{0.93}
\usepackage{amsmath}

\newcommand{\diffs}[3]{\frac{\partial^2 #1}{
\ifx#2#3 
\partial #2^2
\else
\partial #2 \partial #3
\fi
}}

\usepackage{bm}
\newcommand{\sign}{\text{sign}}

\theoremstyle{plain}

\theoremstyle{remark}
\newtheorem{remark}{Remark}
\newtheorem{lemma}{Lemma}
\theoremstyle{remark}
\newtheorem{assumption}{Assumption}

\graphicspath{{figures/}}

\usepackage{tabularx,booktabs}

\begin{document}

\title{Adaptive Dynamic Sliding Mode Control of Soft Continuum Manipulators}
\author{Amirhossein Kazemipour$^{1,2}$, Oliver Fischer$^{1}$, Yasunori Toshimitsu$^{1,3}$, Ki Wan Wong$^{1}$, Robert K. Katzschmann$^{1}$
\thanks{$^{1}$ ETH Zurich, Switzerland}%
\thanks{$^{2}$ The Sapienza University of Rome, Italy}%
\thanks{$^{3}$ The University of Tokyo, Japan}%
\thanks{{\tt\footnotesize \{\href{mailto:akazemi@ethz.ch}{akazemi},\href{mailto:olivefi@ethz.ch}{olivefi},\href{mailto:ytoshimitsu@ethz.ch}{ytoshimitsu},\href{mailto:kiwong@ethz.ch}{kiwong},\href{mailto:rkk@ethz.ch}{rkk}\}@ethz.ch}}
}
\maketitle

\begin{abstract}
Soft robots are made of compliant materials and perform tasks that are challenging for rigid robots. However, their continuum nature makes it difficult to develop model-based control strategies. This work presents a robust model-based control scheme for soft continuum robots. Our dynamic model is based on the Euler-Lagrange approach, but it uses a more accurate description of the robot's inertia and does not include oversimplified assumptions. Based on this model, we introduce an adaptive sliding mode control scheme, which is robust against model parameter uncertainties and unknown input disturbances. We perform a series of experiments with a physical soft continuum arm to evaluate the effectiveness of our controller at tracking task-space trajectory under different payloads. The tracking performance of the controller is around 38\% more accurate than that of a state-of-the-art controller, i.e., the inverse dynamics method. Moreover, the proposed model-based control design is flexible and can be generalized to any continuum robotic arm with an arbitrary number of segments. With this control strategy, soft robotic object manipulation can become more accurate while remaining robust to disturbances.
\end{abstract} 

\vspace{-5pt}
\section{Introduction}
\label{sec:intro}
Soft robotics is a rapidly growing sub-field of robotics. Soft robots are fabricated with compliant and deformable materials, and they can perform tasks that would be extremely challenging for conventional rigid robots~\cite{rus_design_2015}. The inherent compliance of soft manipulators distinguishes them from other types of robots, making them more suitable for interacting with humans and their environment~\cite{qi2017design}. Their continuum properties also enable them to adapt to complex environments in which rigid robots may fail~\cite{laschi2016soft}. 
However, their continuum nature comes at a cost: developing model-based control strategies is complicated.

\begin{figure}[thpb]
\centering
\begin{subfigure}{0.99\columnwidth}
\centering
\includegraphics[height = 4cm, width=0.5\columnwidth]{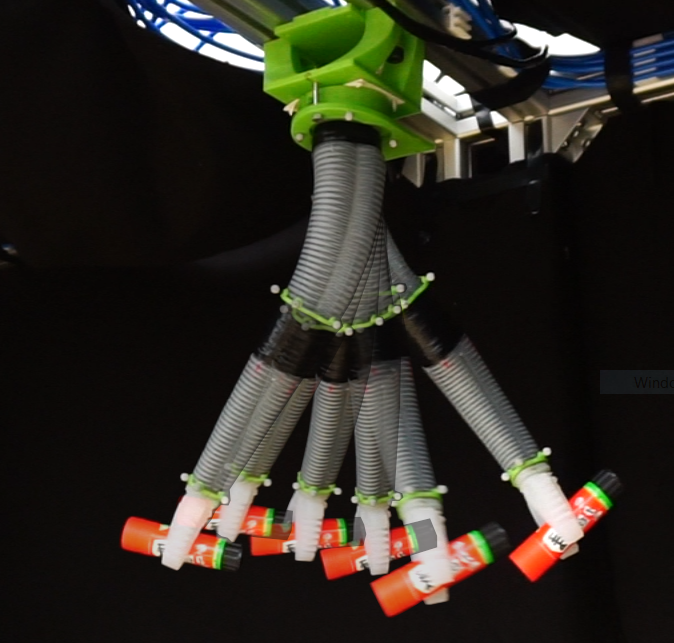}
\vspace{-2pt}
\caption{}%
\vspace{-8pt}
\end{subfigure}
\begin{subfigure}{\columnwidth}
\centering
\includegraphics[width=0.99\columnwidth]{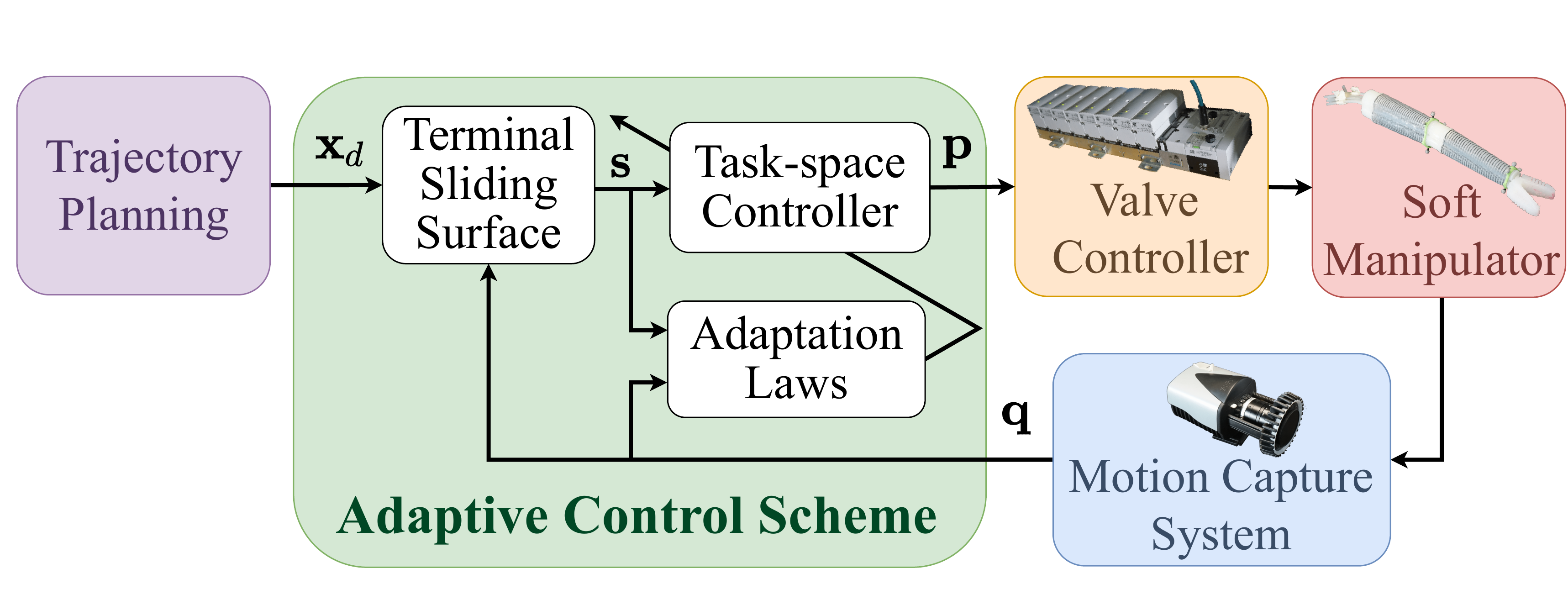}
\vspace{-19pt}
\caption{}%
\label{fig:blk_diagram}
\end{subfigure}\hfill%
\vspace{-4pt}
\caption{Panel (a) shows a dynamically controlled soft robot that is tracking a desired task-space trajectory while carrying a payload with its soft gripper. This robotic arm consists of two segments, and each segment has three chambers that can deform under pressurization. The soft arm is controlled using a model-based adaptive control strategy, as is illustrated in panel (b). This robot is actuated using a proportional valve controller, and the motion capture cameras are used to measure its curvature.}
\label{fig:cover}
\vspace{-25pt}
\end{figure}

There are several approaches to developing a dynamic model for soft robotic arms. For example, a dynamic model can be based on \emph{Koopman Operator theory}~\cite{bruder2019modeling, bruder2020data}, reduced-order finite element models~\cite{katzschmann2019dynamically}, polynomial curvature fitting~\cite{della2019control}, or discrete Cosserat rod models~\cite{till2019real}. In~\cite{katzschmann2019dynamic, della2018dynamic, della2020model}, the dynamic model for the soft continuum arms is based on an approach called \emph{Augmented Rigid Body} formulation. In this approach, the soft robot's motion is approximated with a classic rigid link manipulator, and then it is transformed back into a \emph{Piecewise Constant Curvature} (PCC) formulation for control.
However, as the number of Constant Curvature (CC) segments increases, the auxiliary \emph{rigid states} 
in the model significantly increase the computational burden in the controller loop.
As an alternative, a dynamic model of a soft continuum arm can be derived from the integral Lagrangian approach~\cite{godage2016dynamics} if we assume that the mass distribution along the arm is continuous. This eliminates the need for augmented rigid states in the model. However, when this approach is used to model a continuum arm that has multiple sections, computing the dynamic terms in real-time implementation is computationally inefficient. To address this issue, the mass of each section can be approximated discretely by lumping it into a single mass point along the arm~\cite{falkenhahn2014dynamic, falkenhahn2015model}. However, the lumped-mass models in~\cite{falkenhahn2014dynamic, falkenhahn2015model} assume that the mass of each segment is located at the tip of each segment, and this leads to an inaccurate description of the dynamics, particularly for segments with a high length-to-diameter ratio. Thus, previous models tend to suffer from inefficiencies in the computation of the dynamic parameters or model oversimplifications, resulting in poor performance of model-based controllers.

Prior works on closed-loop control strategies for a soft robotic arm that has to perform dynamic tasks include proportional-derivative (PD) control with dynamic compensation~\cite{katzschmann2019dynamic, della2018dynamic, della2020model} and observer-based dynamic control with a reduced-order finite element model~\cite{katzschmann2019dynamically}. However, these controllers assume perfect knowledge of the model and its parameters.
To make the robot robust to model uncertainties, an adaptive kinematic controller is proposed in~\cite{wang2016visual} and tested on a physical soft robotic arm; however, pure kinematic-based control strategies are not suitable for robots that perform dynamic tasks. To tackle this problem, a popular adaptive control scheme, which was introduced by Li and Slotine~\cite{slotine1988adaptive}, is implemented in curvature-space for a simulated continuum soft arm~\cite{trumic2020adaptive}.
However, this controller is based on the hyper-redundant \emph{Augmented Rigid Body} model. Due to complexities in the Augmented Rigid Body model for a 3D soft manipulator, an adaptive controller is built based on a simplified model that approximates the mass concentrated at the tip of each segment~\cite{della2020improved}. Therefore, previous methods have either not considered the system uncertainties or have been built based on oversimplified models, both of which affect the performance and robustness of the control system.

In this work, we propose a robust model-based control strategy for soft robotic arms. First, the dynamic model is derived from the Lagrangian formulation, without the oversimplified assumption that a mass is concentrated at the segment's tip. On the basis of this model, new adaptive curvature-space and task-space controllers are presented. The proposed control strategy is built upon three main approaches: (1) the well-known Slotine-Li adaptive control scheme~\cite{slotine1987adaptive}, which allows the controller to estimate the dynamic coefficients of the robot online; (2) an adaptive approach to estimating unknown disturbance bounds, which makes the robot robust even when there are unknown input disturbances; and (3) the terminal sliding mode control strategy, which increases the convergence rate of the tracking error, thereby enhancing the overall closed-loop performance.
A series of physical experiments on a real soft robotic arm under various load conditions are used to evaluate the performance and robustness of our model-based adaptive control scheme.
\vspace{-4pt}
\section{Modeling}
\vspace{-4pt}
\label{sec:Modeling}
\subsection{Kinematic model based on Piecewise Constant Curvature}
\label{subsec:Kinematics}
\begin{figure}
\centering
\begin{subfigure}{0.49\columnwidth}
\centering
\includegraphics[height=2.9cm]{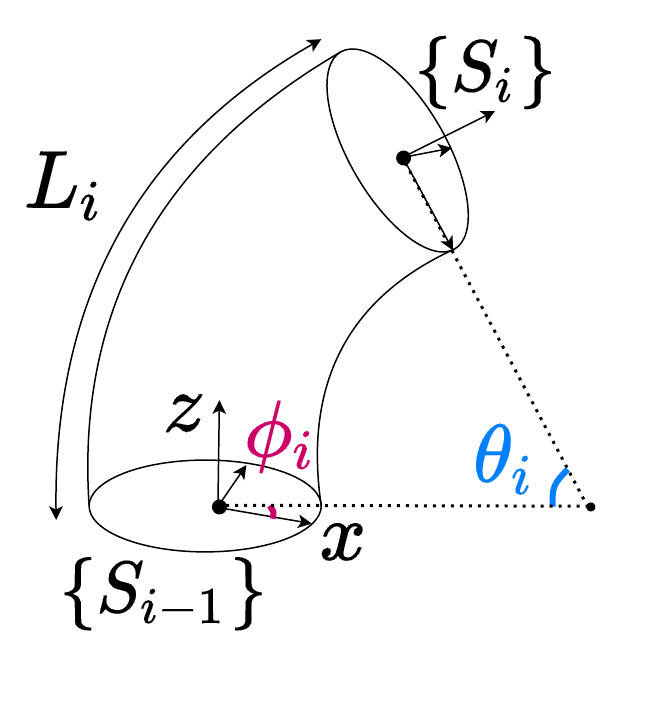}
\vspace{-6pt}
\caption{}%
\label{fig:CC_seg}
\end{subfigure}
\begin{subfigure}{0.49\columnwidth}
\centering
\includegraphics[height=2.9cm]{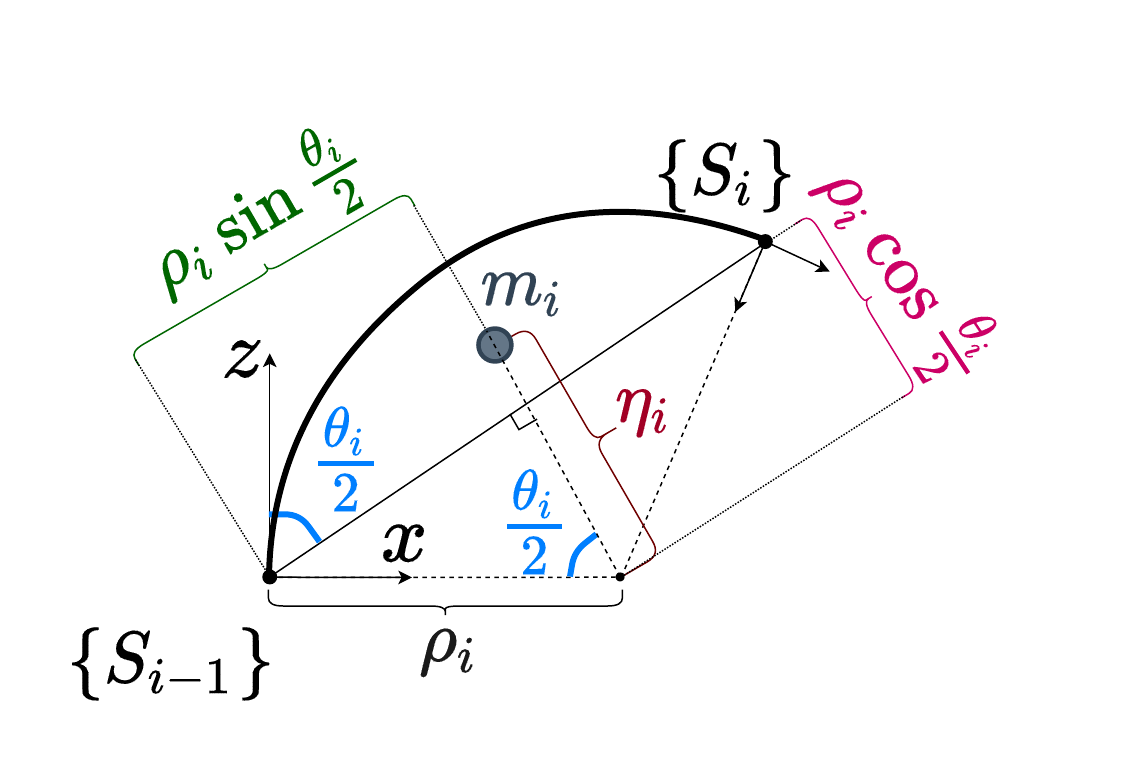}
\vspace{-18pt}
\caption{}%
\label{fig:CoM}
\end{subfigure}\hfill%
\vspace{-3pt}
\caption{(a) shows a CC segment in 3D, and (b) illustrates the CoM position and geometric relationships in a CC segment.\label{fig:kinematic}}
\vspace{-19pt}
\end{figure}

\vspace{-3pt}
To describe the kinematics of the soft robotic arm, we use the PCC approach~\cite{webster2010design}. The base frame $\{S_0\}$ is followed by $n$ reference frames $\{S_1\},\dots,\{S_n\}$, which are attached to the tip of each segment. If we assume that the elongation of the arm's segments is zero, each segment's configuration can be described using only two variables: the angle $\phi$ between the plane $x\,-\,z$ and the plane on which the curvature takes place, and $\theta$, which is the angle of curvature. The curvature radius of each segment satisfies the linear relation $\rho = L^{}/\theta$, where $L$ is the segment's length, which is assumed to be constant. \Cref{fig:CC_seg,fig:CoM} show the kinematic representation of a CC continuum segment. We denote $\bm{q}_i = \begin{pmatrix} \phi_i & \theta_i \end{pmatrix}^T \in \mathbb{R}^{2}$ as the configuration of each segment, and for the whole robot, $\bm{q} \in \mathbb{R}^{2n}$ contains $\bm{q}_i$ for all $n$ segments. We denote $\prescript{i-1}{}{\bm{T}}_i (\phi_i,\theta_i)$ as the transformation between two consecutive reference frames $\{S_{i-1}\}$ and $\{S_i\}$ which can be obtained using geometrical considerations~\cite{katzschmann2019dynamic}.
\vspace{-5pt}
\subsection{Dynamic model based on the Lagrangian approach}
\label{subsec:Dynamics}
\vspace{-3pt}
The Lagrangian is defined as $\mathcal{L} = \mathcal{T}-\mathcal{U}$,
where $\mathcal{T}$ and $\mathcal{U}$ denote the total kinetic and potential energy, respectively~\cite{siciliano2010robotics}. The equations of motion for a robot can be derived from the Lagrangian as $\frac{d}{dt}\,\frac{\partial \mathcal{L}}{\partial \dot{q}_i} -\frac{\partial \mathcal{L}}{\partial q_i} = u_i$, for $i=1,\dots,2n$. 
$u_i$ denotes the generalized force associated with generalized coordinate $q_i \in \mathbb{R}^{2n}$, and $n$ denotes the total number of continuum segments.
\subsubsection{Potential Energy}
The elastic and gravitational potential energies can be computed as  $\mathcal{U}_e = \frac{1}{2} \sum_{i=1}^{n}   k_{s,i} \theta_i^2$ and $\mathcal{U}_g = - \sum_{i=1}^{n} m_i\,\bm{g}_0^T \,\bm{r}_{0,ci} $, respectively. $k_{s,i}$, $m_i$, and $\bm{r}_{0,ci}$ denote the stiffness coefficient, the mass, and the center of mass (CoM) position of the i-th segment, respectively. $\bm{g}_0 = \begin{pmatrix} 0 & 0 & -g\end{pmatrix}^T$ is the gravity acceleration vector in the base reference frame. $g>0$ denotes the gravitational acceleration constant.

\subsubsection{Center of Mass Position}
$\bm{r}_{0,ci}$ denotes the  the i-th segment's CoM position with respect to the robot's base frame $\{S_0\}$. It can be computed as:
\begin{equation} \label{eq:CoM}
\begin{pmatrix} \bm{r}_{0,ci} \\ 1 \end{pmatrix} = \prescript{0}{}{\bm{T}}_1 (\phi_1,\theta_1)\,\dots\,\prescript{i-1}{}{\bm{T}}_i (\phi_i,\theta_i) \begin{pmatrix} \prescript{i}{}{\bm{r}}_{i,ci} \\ 1 \end{pmatrix},
\end{equation}
where $\prescript{i}{}{\bm{r}}_{i,ci}$ is the position of the center of the i-th segment's mass with respect to the $\{S_i\}$ frame. $\prescript{i}{}{\bm{r}}_{i,ci}$ can be expressed using geometrical considerations (see~\cref{fig:CoM}):
\begin{equation}\label{eq:CoM_eachseg}
\prescript{i}{}{\bm{r}}_{i,ci} = \bm{R}_z(\phi_i) \bm{R}_y(\frac{\theta_i}{2}) \begin{pmatrix}
\rho_i \,\cos{\frac{\theta_i}{2}}-\eta_i \\ 0 \\ \rho_i \,\sin{\frac{\theta_i}{2}}
\end{pmatrix},
\end{equation}
where $\rho_i$ is the curvature radius of the i-th segment, and $\eta_i = \frac{2\rho_i \,\sin{\frac{\theta_i}{2}}}{\theta_i}$ is the distance between the CoM and the center of curvature.
\begin{remark}
In contrast to previous models~\cite{falkenhahn2014dynamic,falkenhahn2015model}, which assume that the mass is located at the segment's tip, \cref{eq:CoM_eachseg} considers that the mass is located at each segment's centroid, leading to a more realistic representation of the CoM position.
\end{remark}
\subsubsection{Kinetic Energy}
The total kinetic energy $\mathcal{T}$ can be derived from each segment's individual energy terms as $\mathcal{T} = \sum_{i=1}^{n} \frac{1}{2} m_i\,\bm{v}_{ci}^T\,\bm{v}_{ci}
$, where $m_i$ denotes the mass. The linear velocity of each segment can be computed as $\bm{v}_{ci} = \frac{\partial \bm{r}_{0,ci}}{\partial \bm{q}}\,\dot{\bm{q}}
$. $\bm{r}_{0,ci}$ is obtained in the same way as in \cref{eq:CoM}.
We note that the kinetic energy neglects the contributions of rotational energies, as they are much lower than the translational energies \cite{falkenhahn2015model}. 
\subsubsection{Dynamic Terms}
The dynamic terms can be obtained from the system's potential and kinetic energies~\cite{siciliano2010robotics}. The dynamics model in compact form can be written as:
\begin{equation} \label{eq:dyns}
\bm{M} (\bm{q}) \ddot{\bm{q}} + \bm{C} (\bm{q},\dot{\bm{q}})\dot{\bm{q}} + \bm{D} (\bm{q})\dot{\bm{q}} + \bm{g} (\bm{q}) + \bm{k}(\bm{q}) = \bm{A}(\bm{q}) \bm{p}+\bm{d},
\end{equation}
where $\bm{p} \in \mathbb{R}^{c}$ indicates the air pressure in the robot's fluidic chambers, and the superscript $c$ specifies the total number of chambers. $\bm{M} \in \mathbb{R}^{2n\times2n}$, $\bm{C} \in \mathbb{R}^{2n\times2n}$, $\bm{g} \in \mathbb{R}^{2n\times1}$, $\bm{k} \in \mathbb{R}^{2n\times1}$, and $\bm{A} \in \mathbb{R}^{2n \times c}$ are the generalized inertia matrix, Coriolis/centrifugal matrix, gravity force vector, elastic force vector, and actuator mapping matrix, respectively. $\bm{D} \in \mathbb{R}^{2n\times2n}$ is the damping matrix, which is described in~\cite{katzschmann2019dynamic}. $\bm{d}\in \mathbb{R}^{2n}$ denotes the unknown input disturbances.
\vspace{-5pt}
\subsection{Linear Parameterization}
\label{subsec:SysID}
Neglecting the unknown input disturbances, the dynamic equations in \cref{eq:dyns} can be re-written in a linear form as:
\begin{equation} \label{eq:linear}
\bm{Y} (\bm{q},\dot{\bm{q}},\Ddot{\bm{q}}) \, \bm{a} = \bm{A}(\bm{q}) \bm{p},
\end{equation}
where $\bm{a} \in \mathbb{R}^r$ contains the $r$ number of dynamic coefficients in the robot. $\bm{Y} \in \mathbb{R}^{2n \times r}$ is called the \emph{regression matrix}, which is a known time-varying matrix that depends only on $\bm{q}$, $\dot{\bm{q}}$, and $\Ddot{\bm{q}}$. Vector $\bm{a}$ combines the robot's physical parameters (\textit{i.e.}, the length, mass, stiffness, and damping coefficients of the segments). 
\vspace{-12pt}
\section{Control Synthesis}
\vspace{-10pt}
\label{sec:Controller}
In this section, we present the proposed adaptive control scheme for soft continuum manipulators in both the curvature space and the task space. 
\vspace{-16pt}
\subsection{Adaptive Control: Curvature Space}
\label{subsec:Adaptive_joint}
\vspace{-6pt}
We first begin by modifying the reference trajectories that are presented in~\cite{slotine1987adaptive}. These trajectories allow us to define a terminal sliding manifold. The proposed reference trajectories at velocity and acceleration levels are as follows:
\begin{equation}\label{eq:ref_traj}
 \begin{split}
\dot{\bm{q}}_r &= \dot{\bm{q}}_d + \bm{\Lambda} \sign^{\alpha}(\bm{q}_d - \bm{q}), \\
\ddot{\bm{q}}_r  &= \Ddot{\bm{q}}_d + \alpha \bm{\Lambda} |\bm{q}_d - \bm{q}|^{\alpha-1} (\dot{\bm{q}}_d - \dot{\bm{q}}),
\end{split}   
\end{equation}
where $\sign^{\alpha} (\bm{x}) \triangleq |\bm{x}|^{\alpha}\sign (\bm{x})$. $\bm{q}_d$ is the desired trajectory in the configuration space, $\bm{\Lambda}$ is a constant diagonal matrix with positive diagonal entries, and the exponent $\alpha \in (0.5,1)$ is a constant scalar.
Accordingly, the nonlinear terminal sliding manifold $\bm{s}$ can be defined as $\bm{s} \triangleq \dot{\bm{q}} - \dot{\bm{q}}_r = \dot{\bm{e}}_q + \bm{\Lambda}\sign^{\alpha} (\bm{e}_q)$, where $\bm{e}_q \triangleq \bm{q} - \bm{q}_d$ and $\dot{\bm{e}}_q \triangleq \dot{\bm{q}} - \dot{\bm{q}}_d$.
The control law is presented as follows:
\begin{equation}\label{eq:Jnt_controller}
\begin{split}
\bm{p} &= \bm{A}^{\dagger}(\bm{q})\Big(\hat{\bm{M}}(\bm{q}) \Ddot{\bm{q}}_r + \hat{\bm{C}}(\bm{q},\dot{\bm{q}}) \dot{\bm{q}}_r + \hat{\bm{D}}(\bm{q}) \dot{\bm{q}} + \hat{\bm{g}} (\bm{q}) \\
&+ \hat{\bm{k}}(\bm{q}) - \bm{K}_D \bm{s} - \hat{\bm{b}} \,\sign (\bm{s}) \Big),
\end{split}
\end{equation}
where $\bm{K}_D>0$ is a diagonal matrix, and $\bm{A}^{\dagger}(q)$ is the pseudo-inverse of the mapping matrix.
The estimated feedback-linearization terms in the control law can be computed using the regressor matrix $\bm{Y}$ introduced in \cref{eq:linear} as:
\begin{equation} \label{eq:hat}
\hat{\bm{M}} \Ddot{\bm{q}}_r + \hat{\bm{C}} \dot{\bm{q}}_r +\hat{\bm{D}}\dot{\bm{q}} + \hat{\bm{g}} + \hat{\bm{k}}  = \bm{Y}(\bm{q},\dot{\bm{q}},\dot{\bm{q}}_r,\Ddot{\bm{q}}_r) \hat{\bm{a}}.
\end{equation}
Accordingly, the control law in \cref{eq:Jnt_controller} can be rewritten as:
\begin{equation}\label{eq:Jnt_controller2}
\bm{p} = \bm{A}^{\dagger}(\bm{q}) \left( \bm{Y}(\bm{q},\dot{\bm{q}},\dot{\bm{q}}_r,\Ddot{\bm{q}}_r) \hat{\bm{a}} -\bm{K}_D\bm{s} - \hat{\bm{b}}\,\sign (\bm{s}) \right).
\end{equation}
In the control synthesis, we have considered two adaptation laws:
\begin{equation}\label{eq:adaptationlaw1}
\dot{\hat{\bm{a}}} = -\bm{\Gamma} \bm{Y}^T(\bm{q},\dot{\bm{q}},\dot{\bm{q}}_r,\Ddot{\bm{q}}_r)\, \bm{s},\;\;\;\;\;\;\;\;\dot{\hat{\bm{b}}} = \bm{\Psi} |\bm{s}|.
\end{equation}
The first adaptation law is used to estimate the uncertain dynamic coefficients in vector $\bm{a}$, where $\Hat{\bm{a}}$ is the estimation of dynamic coefficients $\bm{a}$. The second adaptation law is designed to make the control law robust to unknown external disturbances, where $\Hat{\bm{b}}$ is the estimation of the upper bound of disturbances $\bm{b}$. Note that $\bm{\Psi}>0$ and $\bm{\Gamma}>0$ are the diagonal constant gain matrices.

For convenience in the stability proof, we first give the following lemma and assumption:
\begin{lemma}\label{lemma1}
If the Coriolis matrix is defined using Christoffel symbols, the matrix $(\dot{\bm{M}}-2\bm{C})$ is skew-symmetric~\cite{siciliano2010robotics}.
\end{lemma}
\begin{assumption}
The input disturbances $\bm{d}$ are bound by $|d_i| \leq b_i,\; \forall i=1,\dots,n$, where $b_i \in \mathbb{R}$ is an unknown positive scalar.
\label{assumption1}
\end{assumption}
The stability of the control scheme can be proved by considering the following Lyapunov function:
\begin{equation} \label{eq:lyapunov}
V(t) = \frac{1}{2} \bm{s}^T \bm{M} \bm{s} + \frac{1}{2} \Tilde{{\bm{a}}}^T\bm{\Gamma}^{-1}\Tilde{{\bm{a}}} + \frac{1}{2} \Tilde{{\bm{b}}}^T\bm{\Psi}^{-1}\Tilde{{\bm{b}}},
\end{equation}
where $\Tilde{{\bm{a}}} = \bm{a} - \hat{\bm{a}}$ and $\Tilde{{\bm{b}}} = \bm{b} - \hat{\bm{b}}$.
By taking the time-derivative of $V(t)$ along with \cref{eq:dyns}, we obtain:
\setlength{\arraycolsep}{0.0em} \begin{eqnarray} \dot{V} &{}={}& \frac{1}{2} \bm{s}^T \dot{\bm{M}} \bm{s} + \bm{s}^T \bm{M} \dot{\bm{s}} - \Tilde{{\bm{a}}}^T\bm{\Gamma}^{-1}\dot{\hat{{\bm{a}}}} - \Tilde{{\bm{b}}}^T\bm{\Psi}^{-1}\dot{\hat{{\bm{b}}}} \nonumber\\ &&\!\!\!\!\!\!=\frac{1}{2} \bm{s}^T \dot{\bm{M}} \bm{s} + \bm{s}^T \big(\, -\bm{M}\ddot{\bm{q}}_r - \bm{C} \dot{\bm{q}} - \bm{D} \dot{\bm{q}} - \bm{g} \nonumber \\
&&{-}\: \bm{k} + \bm{d} + \bm{A}\bm{p}\,\big)- \Tilde{{\bm{a}}}^T\bm{\Gamma}^{-1}\dot{\hat{{\bm{a}}}} - \Tilde{{\bm{b}}}^T\bm{\Psi}^{-1}\dot{\hat{{\bm{b}}}}. \end{eqnarray} \setlength{\arraycolsep}{5pt}
By replacing the control law in \cref{eq:Jnt_controller} and using \cref{eq:hat}, we have:
\setlength{\arraycolsep}{0.0em} \begin{eqnarray} \dot{V} &{}={}& \frac{1}{2} \bm{s}^T \dot{\bm{M}} \bm{s} + \bm{s}^T \big( -\bm{Y}\Tilde{\bm{a}} - \bm{C}\bm{s} - \bm{K}_D \bm{s} + \bm{d}  -\hat{\bm{b}} \, \sign(\bm{s}) \big) \nonumber\\ &&{-}\:\Tilde{{\bm{a}}}^T\bm{\Gamma}^{-1}\dot{\hat{{\bm{a}}}} - \Tilde{{\bm{b}}}^T\bm{\Psi}^{-1}\dot{\hat{{\bm{b}}}}. \end{eqnarray} \setlength{\arraycolsep}{5pt}
Replacing the adaptation laws in \cref{eq:adaptationlaw1} yields
\setlength{\arraycolsep}{0.0em} \begin{eqnarray} \dot{V} &{}={}& \frac{1}{2} \bm{s}^T (\dot{\bm{M}}-2\bm{C}) \bm{s} - \bm{s}^T \bm{Y}\Tilde{\bm{a}} -\bm{s}^T \bm{K}_D \bm{s} + \bm{s}^T \bm{d} \nonumber\\ &&{-}\:\hat{\bm{b}}^T|\bm{s}| +\Tilde{{\bm{a}}}^T\bm{Y}^T \bm{s}
-(\bm{b}-\hat{\bm{b}})^T |\bm{s}|. \end{eqnarray} \setlength{\arraycolsep}{5pt}
Considering the skew-symmetric property in \cref{lemma1} and the relation $\bm{s}^T\bm{d} \leq |\bm{s}|^T\bm{b}$ followed by assumption~\ref{assumption1}, we obtain $\dot{V} \leq -\bm{s}^T \bm{K}_D \bm{s}$, with $\bm{K}_D>0$.
Therefore, the proposed control law in \cref{eq:Jnt_controller} with the adaptation laws in \cref{eq:adaptationlaw1} force the trajectories to reach the sliding manifold $\bm{s}=0$. When the sliding condition $\bm{s}=0$ is reached, the trajectories are determined by the following differential equation:
\begin{equation} \label{eq:sliding_condition}
\dot{\bm{e}}_q = -\bm{\Lambda} \sign^{\alpha} (\bm{e}_q).
\end{equation}
By directly integrating \cref{eq:sliding_condition}, it can be shown that~\cite{bhat2000finite}, given $\bm{e}_q(0)\neq 0$, the trajectories will reach $\bm{e}_q=0$ in a finite time. This can be determined by $t_{f,i} = \Lambda_i^{-1}(1-\alpha)^{-1}|e_{q,i}(0)|^{1-\alpha}$.
Therefore, $\bm{e}_q=0$ is a terminal attractor (i.e., the tracking errors converge to zero in a finite time).

To avoid overestimating the dynamic coefficients, we use the boundary layer technique in \cite{slotine1983tracking,wang2017adaptive} by defining a new variable as ${s}_{\Delta\,i} = {s}_i - \phi_i\,\text{sat}(\frac{{s}_i}{\phi_i}),$ for $i=1,\dots,n$, where $\phi_i>0$ is the boundary layer thickness. The saturation function is defined as:
\begin{equation} \label{eq:saturation}
\text{sat} (\frac{s_i}{\phi_i}) = \begin{cases}
            \text{sign} (s_i) & \text{if } |s_i|\geq \phi_i \\
            \frac{s_i}{\phi_i} & \text{if } |s_i|< \phi_i
        \end{cases}.
\end{equation}
Inside the boundary layer ($|s_i|< \phi_i$), the new variable ${s}_{\Delta\,i}$ features ${s}_{\Delta\,i} = 0$, while outside the boundary layer ($|s_i|\geq \phi_i$), the relation $\dot{s}_{\Delta\,i} = \dot{s}_i$ is satisfied.
Accordingly, the adaptation laws are modified as:
\begin{equation}
\dot{\hat{\bm{a}}} = -\bm{\Gamma} \bm{Y}^T(\bm{q},\dot{\bm{q}},\dot{\bm{q}}_r,\Ddot{\bm{q}}_r)\, \bm{s}_\Delta,\;\;\;\;\;\;\dot{\hat{\bm{b}}} = \bm{\Psi} |\bm{s}_\Delta|.
\end{equation}
The saturation function defined in~\cref{eq:saturation} can also be used to replace the sign function in~\cref{eq:Jnt_controller}, thereby eliminating the chattering phenomenon caused by the switching function.
\begin{remark}
Compared to the adaptive controller in \cite{trumic2020adaptive}, the control law presented here has an additional term $\hat{\bm{b}}\sign(\bm{s})$. As shown in the stability proof, this term can reject the bounded input disturbances. Moreover, thanks to the second adaptation law in \cref{eq:adaptationlaw1}, there is no need to have prior knowledge of disturbances; the controller rejects them by adjusting the adaptive gains in the switching term. 
\end{remark}
\begin{remark}
By choosing the reference trajectories as in \cref{eq:ref_traj}, the resulting sliding surface ($\bm{s}$) becomes a \emph{Terminal Sliding Manifold} (TSM). Note that, for $\alpha=1$, $\bm{s}$ is equivalent to the conventional linear sliding surface, and \cref{eq:sliding_condition} becomes $\dot{\bm{e}}_q=-\bm{\Lambda}\bm{e}_q$. TSM is widely used in various applications~\cite{kazemipour2017adaptive,kazemipour2020adaptive,yu2020terminal} because it guarantees the convergence of tracking errors on the sliding manifold in finite-time~\cite{yu2020terminal}. This is faster than the asymptotic convergence of the classic linear sliding surface, which is used in Slotine-Li's adaptive method~\cite{slotine1987adaptive}.
\end{remark}
\vspace{-10pt}
\subsection{Adaptive Control: Task Space}
\label{sec:Controller_taskspace}
\vspace{-5pt}
The adaptive controller in~\cref{subsec:Adaptive_joint} can be extended to the task space by replacing the reference trajectories in \cref{eq:ref_traj} with the following:
\begin{equation}
\begin{split}
\dot{\bm{q}}_r &= \bm{J}^{\dagger} \left( \dot{\bm{x}}_d + \bm{\Lambda} \sign^\alpha (\bm{x}_d - \bm{x}) \right),\\
\ddot{\bm{q}}_r &= \bm{J}^{\dagger} \left( \ddot{\bm{x}}_d + \alpha \bm{\Lambda} |\bm{x}_d - \bm{x}|^{\alpha-1} (\dot{\bm{x}}_d - \dot{\bm{x}}) - \dot{\bm{J}} \dot{\bm{q}}_r\right),
\end{split}
\end{equation}
where $\bm{J}^{\dagger}$ stands for the pseudo-inverse of the Jacobian matrix $\bm{J}$. Accordingly, we have
\begin{equation}
\bm{s} = \dot{\bm{q}} - \dot{\bm{q}}_r = \bm{J}^{\dagger} \left( \bm{J}\dot{\bm{q}} -  \dot{\bm{x}}_d + \bm{\Lambda} \sign^\alpha (\bm{x} - \bm{x}_d) \right).
\end{equation}
The structure of the controller and the adaptation laws are the same as in \cref{eq:Jnt_controller,eq:adaptationlaw1}.
For the stability proof, the same Lyapunov function as in \cref{eq:lyapunov} can be used. This results in $\dot{V} \leq  -\Bar{\bm{s}}^T\bm{H}\Bar{\bm{s}}
$, where $\Bar{\bm{s}} \triangleq \bm{J}\dot{\bm{q}} -  \dot{\bm{x}}_d + \bm{\Lambda} \sign^\alpha (\bm{x} - \bm{x}_d)$ and $\bm{H} \triangleq \bm{J}^{\dagger^T}\bm{K}_D\bm{J}^{\dagger}>0$ are used.
Therefore, the trajectories are guaranteed to reach $\bar{\bm{s}}=0$. When the condition $\bar{\bm{s}}=0$ is satisfied, the trajectories are determined via the following differential equation:
\begin{equation} \label{eq:sliding_condition2}
\bm{J}\dot{\bm{q}} -  \dot{\bm{x}}_d + \bm{\Lambda} \sign^\alpha (\bm{x} - \bm{x}_d)=0.
\end{equation}
Let $\bm{e}_x \triangleq \bm{x}-\bm{x}_d$ and $\dot{\bm{e}}_x \triangleq \dot{\bm{x}}-\dot{\bm{x}}_d$ denote the errors, replacing the kinematic relation $\dot{\bm{x}} = \bm{J}\dot{\bm{q}}$ into \cref{eq:sliding_condition2} yields:
\begin{equation} \label{eq:sliding_condition3}
\dot{\bm{e}}_x = -\bm{\Lambda}\sign^\alpha (\bm{e}_x).
\end{equation}
The differential equation in \cref{eq:sliding_condition3} is similar to the one in \cref{eq:sliding_condition}. Thus, the finite time convergence of Cartesian tracking errors on the sliding manifold is guaranteed.
\vspace{-5pt}
\section{Results and Discussions}
\label{sec:results}
\vspace{-5pt}
\subsection{Experimental Setup}
\label{subsec:ExpSteup}
\begin{figure}[htpb]
    \centering
    \includegraphics[width=0.99\columnwidth]{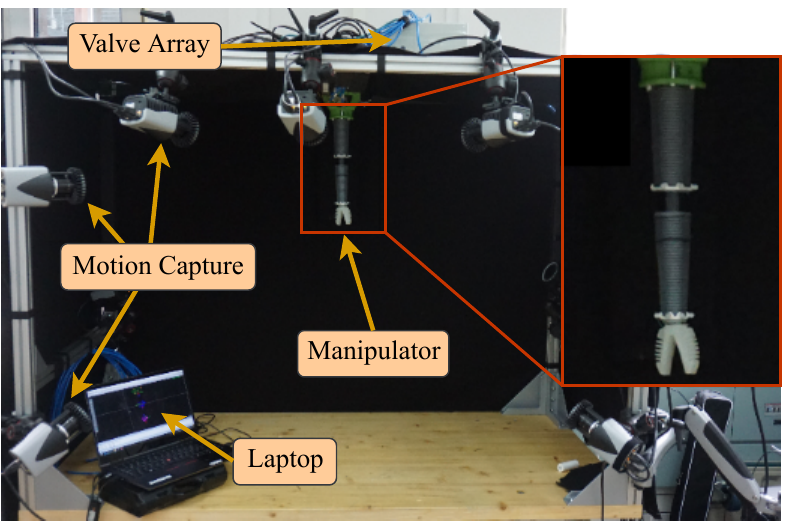}
    \vspace{-12pt}
    \caption{We used a fiber-reinforced soft arm as the experimental platform for validation. The arm consists of two segments with interior pressure chambers. The proportional valve array independently pressurizes the chambers. Motion capture cameras measure the curvature of the segments.\label{fig:expsetup}}
    \vspace{-22pt}
\end{figure}

\vspace{-5pt}
The experimental setup, including the soft robotic arm, is shown in~\cref{fig:expsetup}. It consists of a soft arm with a gripper, a proportional valve manifold, and a motion capture system~\cite{toshimitsu2021sopra}. The arm's length
is $\SI{27}{\cm}$ and it weighs $\SI{276}{\gram}$ in total. The arm has three inflatable chambers in each of its two segments. It is made of silicone elastomer and reinforced with fibers to reduce bloating and increase bending under pressure. The soft gripper and each of the six chambers were actuated independently through an array of proportional valves. The robot configurations were measured in real-time with a motion capture system that consisted of eight infrared (IR) cameras. These cameras were mounted around the arm and connected to a laptop that was running a motion capture software. The reflective markers were attached to the robot base and around the tip of each segment. The valve system can provide up to 2 bars of pressure, but, for our experiments, we never exceeded $\SI{1.2}{\bar}$ to avoid rupturing the arm.
\vspace{-5pt}
\subsection{Experimental Validation Results}
\vspace{-2pt}
\subsubsection{Model Validation}
\label{subsec:SysIDResults}
To validate the dynamic model, we used the dynamic coefficients that are described in \cref{eq:dynamic_coef2}. We chose to define the vector of dynamic coefficients as $\bm{a} = \begin{pmatrix} \bm{a}_1^T & \bm{a}_2^T \end{pmatrix}^T \in \mathbb{R}^{11 \times 1}
$,
where 
\begin{equation*}
\arraycolsep=2.4pt\def\arraystretch{1}
\bm{a}_1^T \triangleq \left(\!\begin{array}{cccccccc}
m_1 L_1^2 & m_2 L_1^2 & m_2 L_2^2 & m_2 L_1 L_2 & m_1 L_1 & m_2 L_1 & m_2 L_2
\end{array}\!\right)
\label{eq:augment_mat}
\end{equation*}
\begin{equation}\label{eq:dynamic_coef2}
\bm{a}_2^T \triangleq \begin{pmatrix}
k_{s,1} & k_{s,2} & k_{d,1} & k_{d,2}
\end{pmatrix}.
\end{equation}
Thus, the vector $\bm{a}_1$, which only contains the mass and length of each segment, could be directly measured, but we carried out a data acquisition procedure to estimate the dynamic coefficients in $\bm{a}_2$. We collected identification data for nine experiments. In each experiment, we injected a sinusoidal input under pressure into fully pneumatic valves. The amplitudes of the sinusoidal inputs were $\SI{0.4}{\bar}$, $\SI{0.6}{\bar}$, and $\SI{0.8}{\bar}$. The periods were $\SI{8}{\second}$, $\SI{16}{\second}$, and $\SI{24}{\second}$. Given that the factorization of the regressor matrix is $\begin{pmatrix}
\bm{Y}_1 & \bm{Y}_2
\end{pmatrix} \begin{pmatrix}
\bm{a}_1^T & \bm{a}_2^T
\end{pmatrix}^T = \bm{A} \bm{p}$ , the regression problem can be defined as $\bm{Y}_2 {\bm{a}}_2 = \left( \bm{A}\bm{p} - \bm{Y}_1 {\bm{a}}_1 \right)$. This can be solved by the pseudo-inverse approach~\cite{gaz2019dynamic}. The estimated values for the stiffness and damping coefficients, along with the mass and length of each segment, are reported in \cref{tb:physical_charc}.
\begin{table}[thpb]
\centering
\vspace{-8pt}
\caption{The physical characteristics of the arm.}
\vspace{-5pt}
\begin{tabular}{@{}ccccc@{}}
\toprule
Segment \# & $m\;[\SI{}{\gram}]$ & $L\;[\SI{}{\cm}]$ & $\hat{k}_s\;[\SI{}{\newton\meter}]$ & $\hat{k}_d\;[\SI{}{\newton\meter\second}]$ \\ \midrule
1       & 154              & 13.5           & 0.124                            & 0.011                                   \\
2       & 122              & 13.5           & 0.083                            & 0.009                                   \\ \bottomrule
\end{tabular}\label{tb:physical_charc}
\vspace{-3pt}
\end{table}

\begin{figure}[htpb]
\vspace{-7pt}
\centering
\begin{subfigure}{0.49\columnwidth}
\centering
\includegraphics[width=\columnwidth]{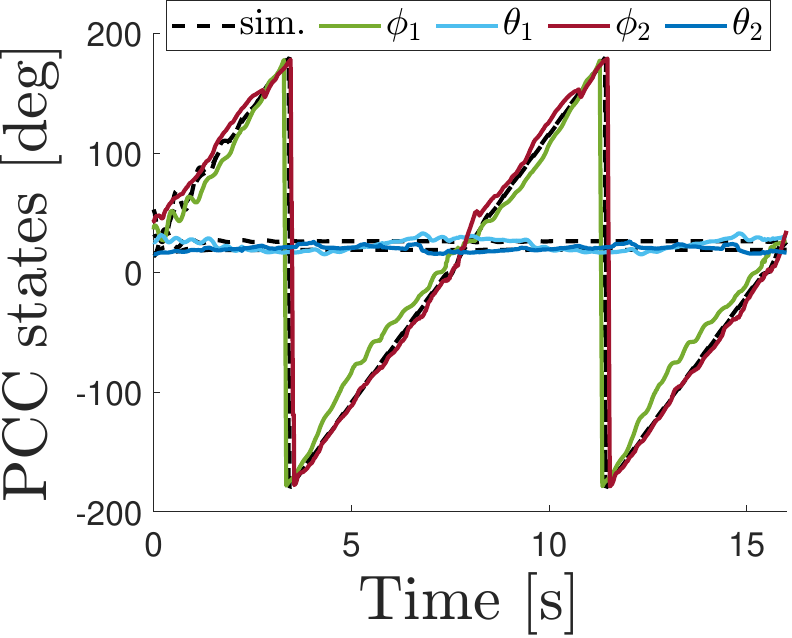}
\vspace{-17pt}
\caption{}%
\label{fig:sysID}
\end{subfigure}\hspace{0.1mm}%
\begin{subfigure}{0.49\columnwidth}
\centering
\includegraphics[width=\columnwidth]{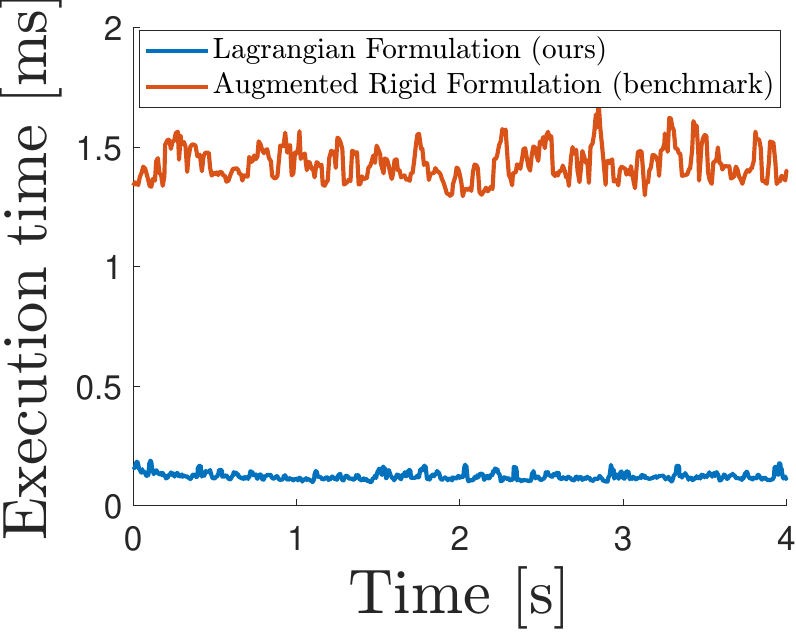}
\vspace{-17pt}
\caption{}%
\label{fig:time_exc}
\end{subfigure}\hfill%
\vspace{-5pt}
\caption{(a) The evolution of PCC parameters, which were obtained from a numerical simulation of the model and from experimental results. The robot was actuated with pre-defined pressure profiles. (b) A comparison of the computational cost of the \emph{Augmented Rigid Body} model versus that of our Lagrangian-based model. The execution time for updating the dynamic terms in real-time for each control cycle is shown in ms.}
\end{figure}

\begin{figure}[thpb]
\vspace{-12pt}
    \centering
    \includegraphics[width=0.99\columnwidth]{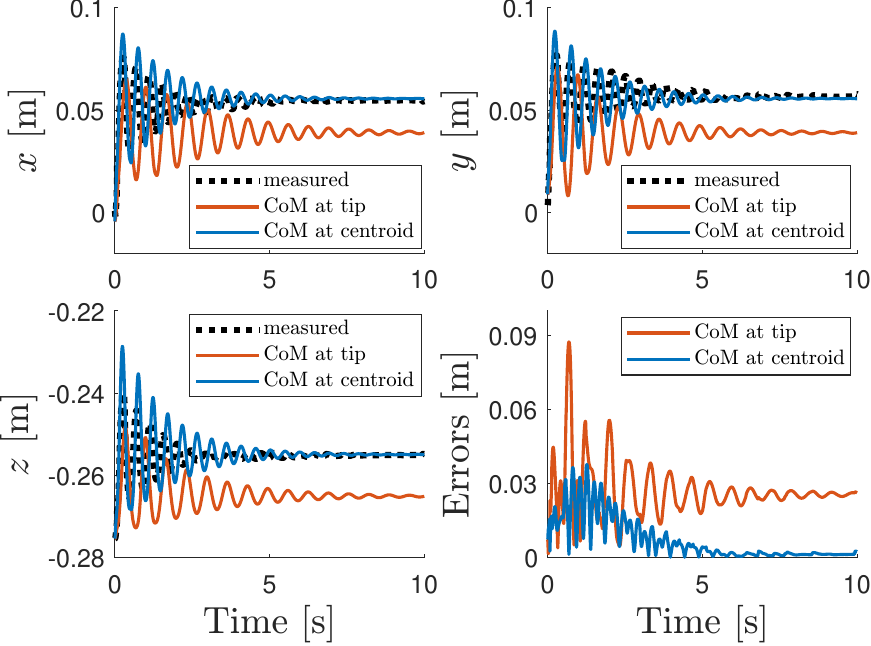}
    \vspace{-18pt}
    \caption{The tip coordinate trajectories and position errors for a step response. The dotted black lines are the experimental trajectories that were measured by the motion capture system. The red lines are the simulation results for the dynamic model in which the CoM is considered to be at the tip of each segment (as in \cite{falkenhahn2014dynamic,falkenhahn2015model}). The blue lines are the simulation results for our dynamic model in which the CoM is considered to be at the centroid (as presented in \cref{eq:CoM_eachseg}).
    \label{fig:CoM_Comparison}}
    \vspace{-25pt} 
\end{figure}

After estimating the numerical values for the dynamic coefficients, we validated our model by actuating the robot through pre-defined feed-forward pressures. We then compared the results of the simulated robot with the results of the real robot. To do this, the chambers of each segment were actuated as $p_i (t) = A\sin^2{\left(\frac{2\pi t}{T} + i\frac{2\pi}{3}\right)},\;\;\;\;i=0,1,2
$,
where $i$ denotes the index of the chambers for each segment. $A=\SI{0.4}{\bar}$ and $T=\SI{16}{\second}$. As is shown in \cref{fig:sysID}, the evolutions of $\phi_i$ and $\theta_i$ over time for the simulated robot closely match with the experimental results.
To verify the efficacy of our model, we compared the step response of the dynamic model in which the center of mass is considered to be at the centroid of each segment (as is described in \cref{eq:CoM_eachseg}) with the step response of dynamic model in which the center of mass is located at the tip of each segment (as is described in \cite{falkenhahn2014dynamic,falkenhahn2015model}). The amplitude of the step input under pressure was $\bm{p}= \begin{pmatrix}
0.4 & 0.3 & 0 & 1.1 & 0.8 & 0
\end{pmatrix}^T\,\si{\bar}$. \Cref{fig:CoM_Comparison} shows the resulting trajectories of the robot's end-effector and the position error, which is the Euclidean distance between the simulated and measured position. The simulated response of our model closely matches the experimental data, while the dynamic model in \cite{falkenhahn2014dynamic,falkenhahn2015model} has a steady-state error of around $\SI{3}{cm}$.
\subsubsection{Task-space Control}
The effectiveness of the proposed model-based control scheme in the task-space was demonstrated on a physical soft robotic arm. To implement our adaptive controller, we first developed the dynamic model in explicit form using the Lagrangian-based approach that we discussed in~\cref{subsec:Dynamics}. The dynamic coefficients vector $\bm{a}$ (see \cref{eq:dynamic_coef2}) was then used to achieve the linear parameterization that is described in \cref{eq:linear}. This approach enabled us to generate a \emph{C++} function for the regressor matrix $\bm{Y}$, which was used in the control loop. 
To benchmark the task-space control, we used an inverse dynamics controller that is similar to the one in~\cite{katzschmann2019dynamic}. For the benchmark controller, we used the \emph{Augmented Rigid Body} dynamic formulation that is described in \cite{katzschmann2019dynamic}. The robot motion library \emph{Drake}~\cite{drake} was used for this framework to calculate the \emph{rigid states} and the corresponding dynamic terms of the rigid-equivalent model. The gains of our adaptive controller were $\Lambda_i = 6.3$, $\alpha = 0.75$, and $K_{D,i} = 0.03$, and the gains of the benchmark controller were $K_p = 12$ and $K_d=4.5$. We have considered two types of Cartesian reference trajectories, each of which has two different sets of timing parameters. The first trajectory was a circle with radius of $\SI{12}{\cm}$ in the $xy$ plane. The second was a 3D star-shaped path in which the vertices were inscribed in a circle with radius of $\SI{14}{\cm}$ in the $xy$ plane. The motion timing law was determined using a trapezoidal velocity profile with different maximum velocity/acceleration parameters for slow and fast trajectories.
Moreover, to show the robustness of the controller, we loaded different payload masses onto the robotic gripper at the end-effector. The controllers did not have information about the payload mass \emph{a priori}.

\begin{figure*}[thpb]
\centering
\begin{subfigure}{.33\textwidth}
\includegraphics[width=\columnwidth]{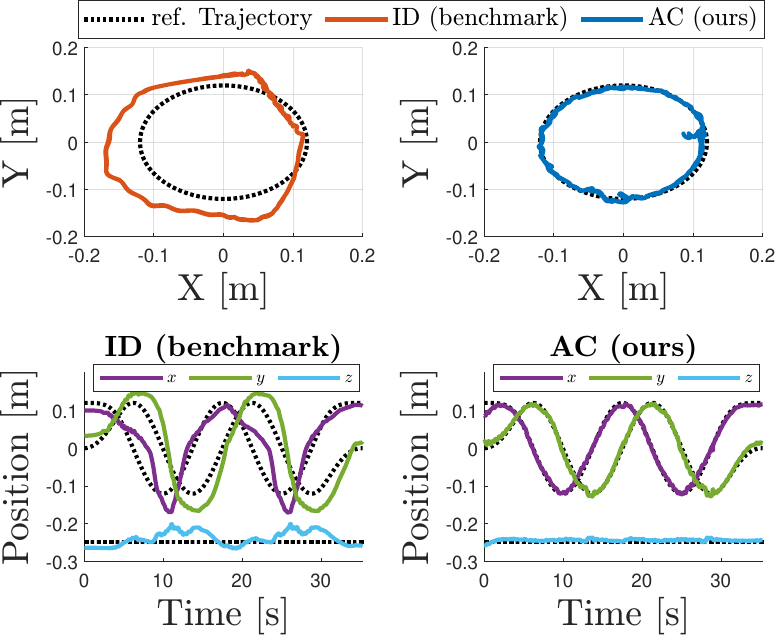}%
\caption{unloaded case}%
\end{subfigure}\hfill%
\begin{subfigure}{.33\textwidth}
\includegraphics[width=\columnwidth]{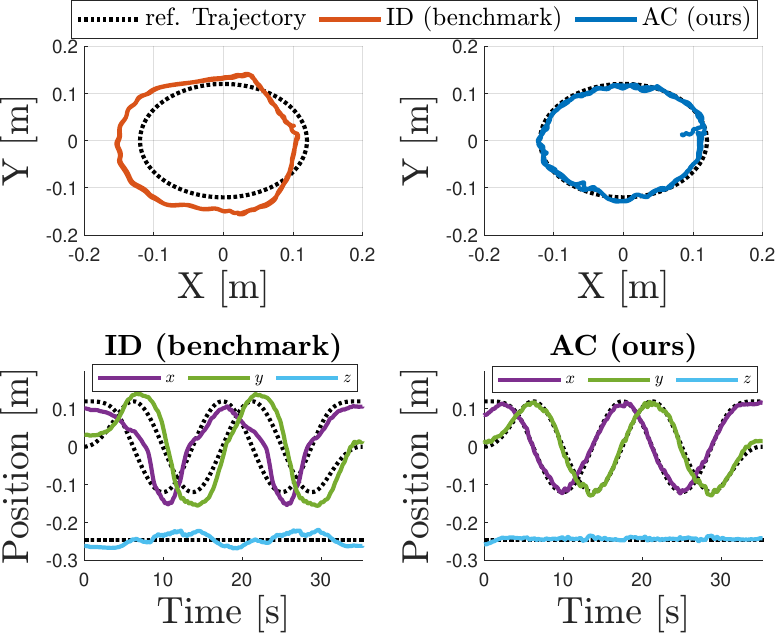}%
\caption{$\SI{12}{\gram}$ payload}%
\end{subfigure}\hfill%
\begin{subfigure}{.33\textwidth}
\includegraphics[width=\columnwidth]{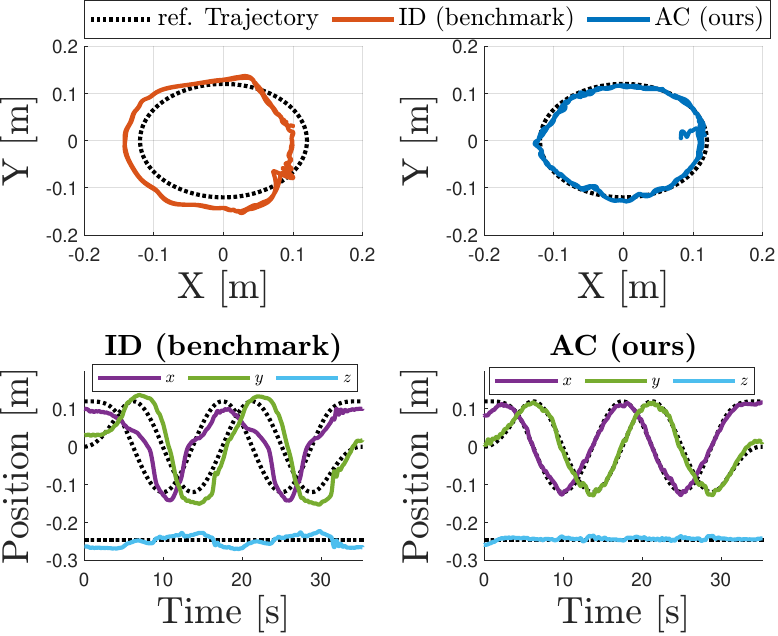}%
\caption{$\SI{25}{\gram}$ payload}%
\end{subfigure}%
\hfill%
\begin{subfigure}{.33\textwidth}
\includegraphics[width=\columnwidth]{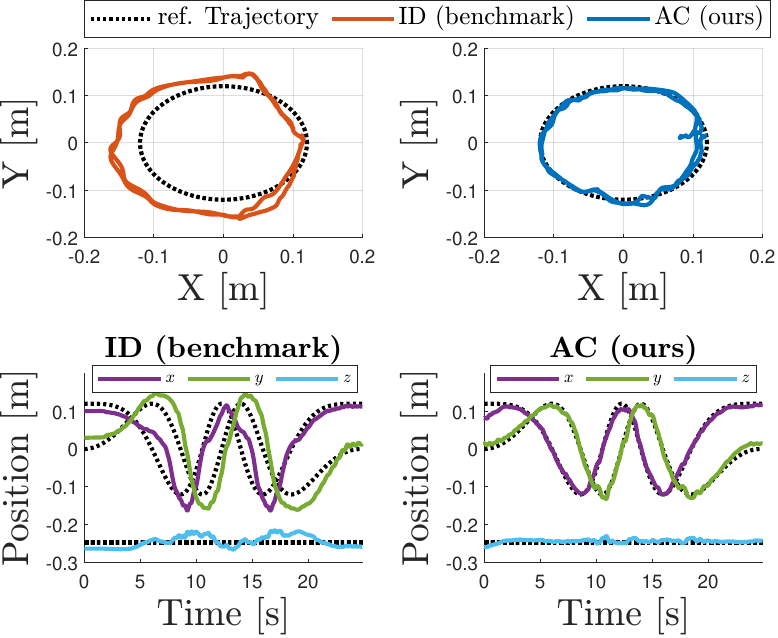}%
\caption{unloaded case}%
\end{subfigure}\hfill%
\begin{subfigure}{.33\textwidth}
\includegraphics[width=\columnwidth]{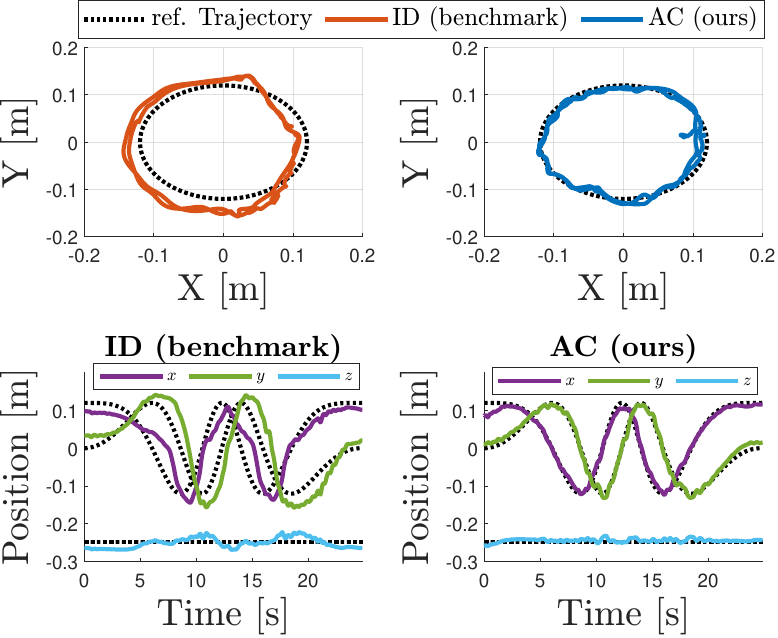}%
\caption{$\SI{12}{\gram}$ payload}%
\end{subfigure}\hfill%
\begin{subfigure}{.33\textwidth}
\includegraphics[width=\columnwidth]{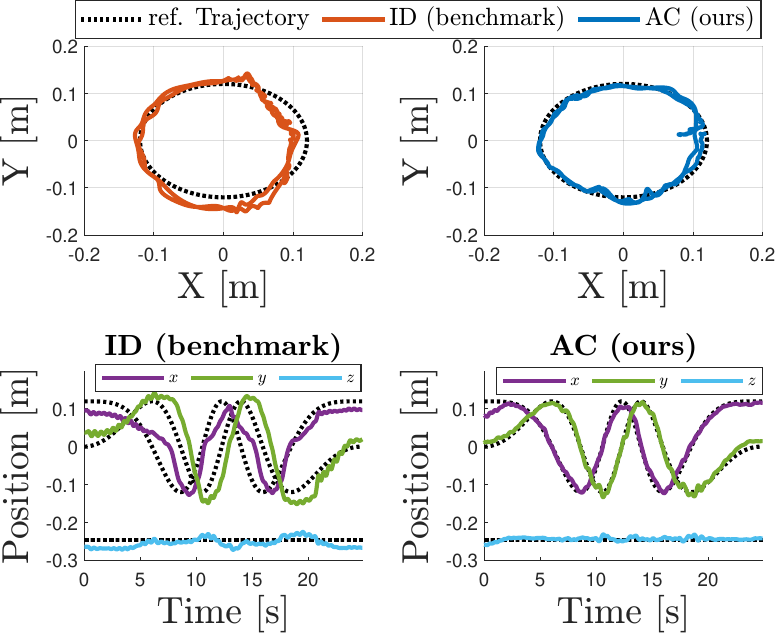}%
\caption{$\SI{25}{\gram}$ payload}%
\end{subfigure}%
\caption{The experimental tracking results for a circular reference trajectory (dotted black line) using our adaptive controller (blue line) and the benchmark controller (red line), both of which carried different payloads. In (a), (b), and (c), the maximum reference velocity and acceleration were set to $v_{max} = \SI{0.05}{\meter\per\second}$ and $a_{max} = \SI{0.01}{\meter\per\second\squared}$, respectively. In (d), (e), and (f), we use $v_{max} = \SI{0.11}{\meter\per\second}$ and $a_{max} = \SI{0.01}{\meter\per\second\squared}$ in the timing law.\label{fig:circular_traj}}
\vspace{-20pt}
\end{figure*}

A comparison of our controller performance against the benchmark is shown in a video\footnote{Video of the real-world experiments including Cartesian trajectory tracking: \url{https://www.youtube.com/watch?v=os5SuStpqh8}} of the experiments.
\Cref{fig:circular_traj} and \Cref{fig:star_traj} show the quantitative results that we obtained for the circular and star-shaped trajectories, respectively. The quantitative performance comparisons of our adaptive controller (AC) and the benchmark controller (ID) for various trajectories are reported in \cref{tb:performance_comp}. The error is defined as the Euclidean distance between the desired position and the measured position. In \cref{tb:performance_comp}, $M_e$, $\mu_e$, and $\sigma_e$ denote the maximum absolute value, average, and standard deviation of the tracking errors, respectively. 'C' and 'S' stand for the circular and star-shaped trajectories, respectively. The low and high reference velocities correspond to different velocity/acceleration values used in the timing law. When the robot is loaded with a payload and when faster motions are considered, the performance of the inverse dynamics controller degrades significantly. However, our proposed adaptive controller achieves higher robustness in handling payload mass variations and maintains a relatively similar performance at both slow and fast reference trajectories.

In~\cref{fig:time_exc}, we compare our Lagrangian-based modeling approach and the \emph{Augmented Rigid Body} modeling technique in terms of computational efficiency; namely, the execution time that was required to update the dynamic terms. The \emph{Augmented Rigid Body} modeling method was implemented via the \emph{Drake} \emph{C++} library \cite{katzschmann2019dynamic}. As is illustrated in~\cref{fig:time_exc}, our Lagrangian-based model updated the dynamic terms around one order of magnitude faster than the \emph{Augmented Rigid Body} model.

\begin{table}[!t]
\renewcommand{\arraystretch}{0.93}
\centering
\caption{Performance indices of the experiments.}
\vspace{-5pt}
\begin{tabular}{@{}ccccccccc@{}}
\toprule
  &    &    & \multicolumn{3}{c}{Low Ref. Velocity} & \multicolumn{3}{c}{High Ref. Velocity} \\ \midrule
\begin{tabular}[c]{@{}c@{}}Ref.\\ Traj.\end{tabular} &
  \begin{tabular}[c]{@{}c@{}}Cont.\\ Type\end{tabular} &
  \begin{tabular}[c]{@{}c@{}}Load\\ {[}g{]}\end{tabular} &
  \begin{tabular}[c]{@{}c@{}}$M_e$\\ {[}cm{]}\end{tabular} &
  \begin{tabular}[c]{@{}c@{}}$\mu_e$\\ {[}cm{]}\end{tabular} &
  \begin{tabular}[c]{@{}c@{}}$\sigma_e$\\ {[}cm{]}\end{tabular} &
  \begin{tabular}[c]{@{}c@{}}$M_e$\\ {[}cm{]}\end{tabular} &
  \begin{tabular}[c]{@{}c@{}}$\mu_e$\\ {[}cm{]}\end{tabular} &
  \begin{tabular}[c]{@{}c@{}}$\sigma_e$\\ {[}cm{]}\end{tabular} \\ \midrule
C & ID & 0  & 10.45       & 5.23       & 2.23       & 9.26         & 4.38       & 1.94       \\
C & AC & 0  & 3.98        & 0.99       & 0.47       & 4.24         & 1.22       & 0.62       \\
C & ID & 12 & 7.87        & 4.14       & 1.39       & 10.95        & 4.86       & 2.19       \\
C & AC & 12 & 4.02        & 1.16       & 0.58       & 4.22         & 1.27       & 0.61       \\
C & ID & 25 & 8.67        & 4.14       & 1.52       & 11.31        & 5.03       & 2.30       \\
C & AC & 25 & 4.24        & 1.21       & 0.61       & 4.43         & 1.27       & 0.67       \\
S & ID & 0  & 6.25        & 3.65       & 1.47       & 10.15        & 4.65       & 2.69       \\
S & AC & 0  & 7.41        & 2.36       & 1.17       & 11.50        & 3.98       & 2.57       \\
S & ID & 12 & 7.22        & 3.76       & 1.73       & 10.69        & 5.31       & 2.81       \\
S & AC & 12 & 5.51        & 2.62       & 1.13       & 9.96         & 4.31       & 2.47       \\
S & ID & 25 & 7.16        & 3.83       & 1.77       & 12.10        & 5.39       & 3.06       \\
S & AC & 25 & 6.80        & 2.93       & 1.21       & 9.67         & 3.83       & 2.31       \\ \bottomrule
\end{tabular}\label{tb:performance_comp}
\vspace{-22pt}
\end{table}

\newcommand\custheight{4.5cm}

\begin{figure*}[thpb]
\centering
\begin{subfigure}{.67\columnwidth}
\includegraphics[height = \custheight]{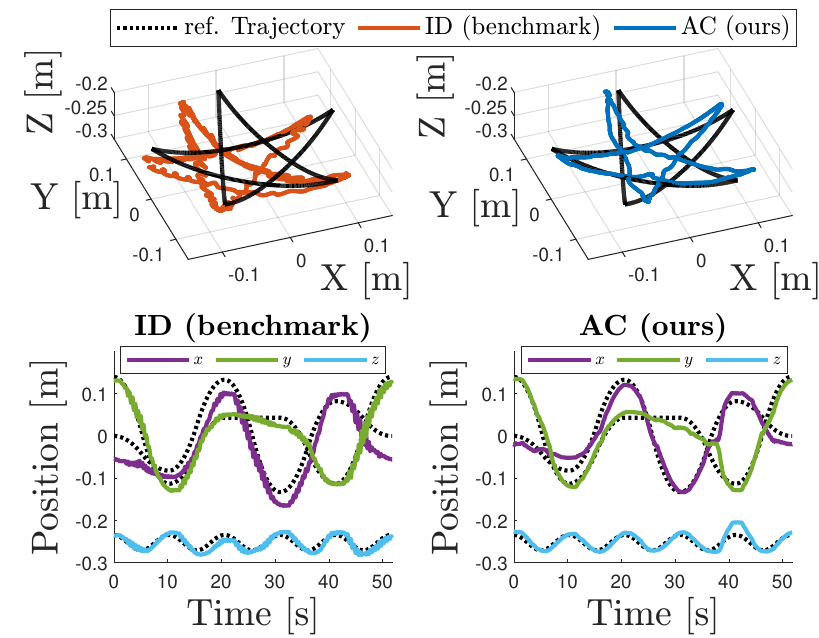}%
\caption{unloaded case}%
\end{subfigure}\hfill%
\begin{subfigure}{.67\columnwidth}
\includegraphics[height = \custheight]{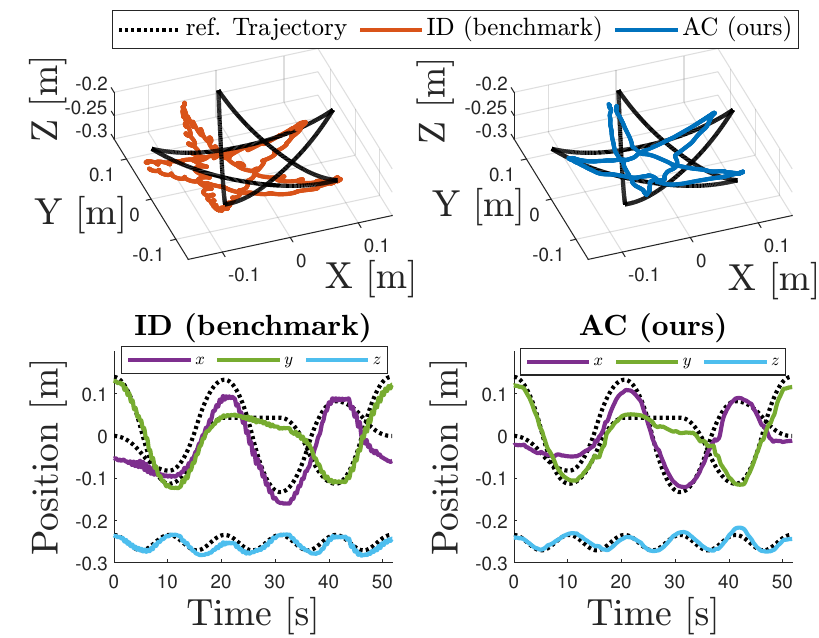}%
\caption{$\SI{12}{\gram}$ payload}%
\end{subfigure}\hfill%
\begin{subfigure}{.67\columnwidth}
\includegraphics[height = \custheight]{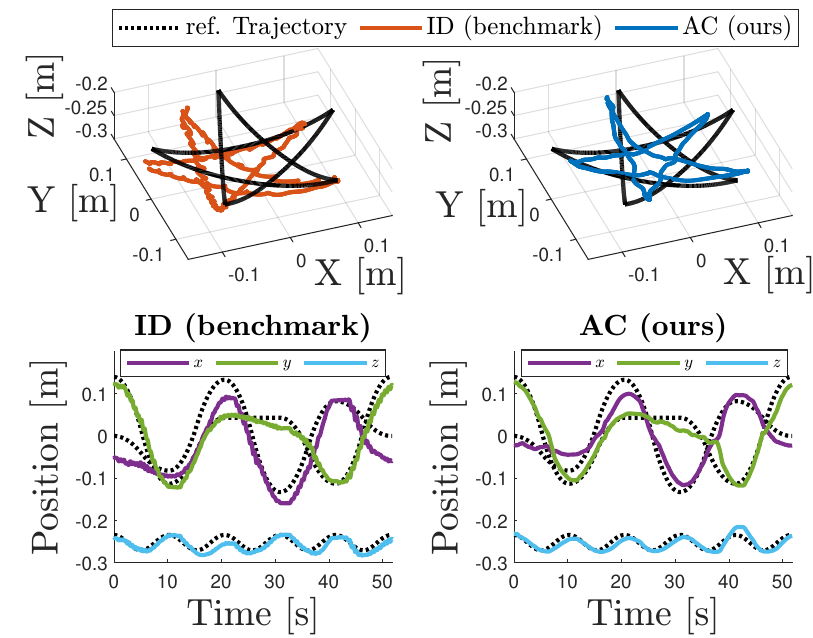}%
\caption{$\SI{25}{\gram}$ payload}%
\end{subfigure}%
\hfill%
\begin{subfigure}{.67\columnwidth}
\includegraphics[height = \custheight]{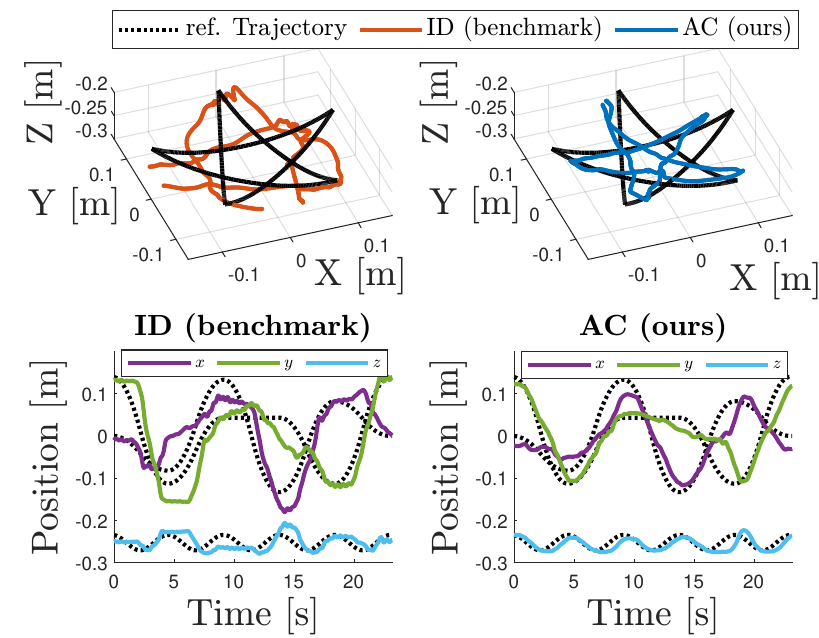}%
\caption{unloaded case}%
\end{subfigure}\hfill%
\begin{subfigure}{.67\columnwidth}
\includegraphics[height = \custheight]{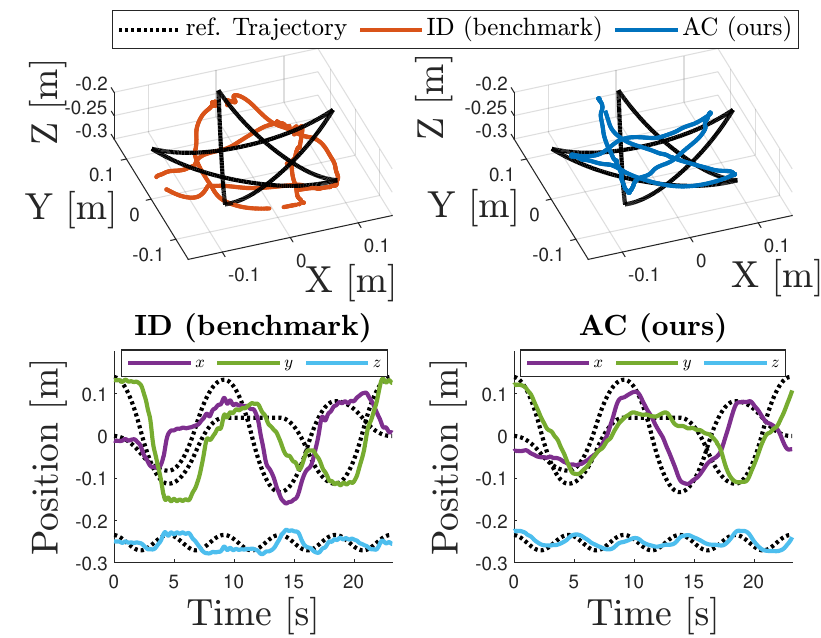}%
\caption{$\SI{12}{\gram}$ payload}%
\end{subfigure}\hfill%
\begin{subfigure}{.67\columnwidth}
\includegraphics[height = \custheight]{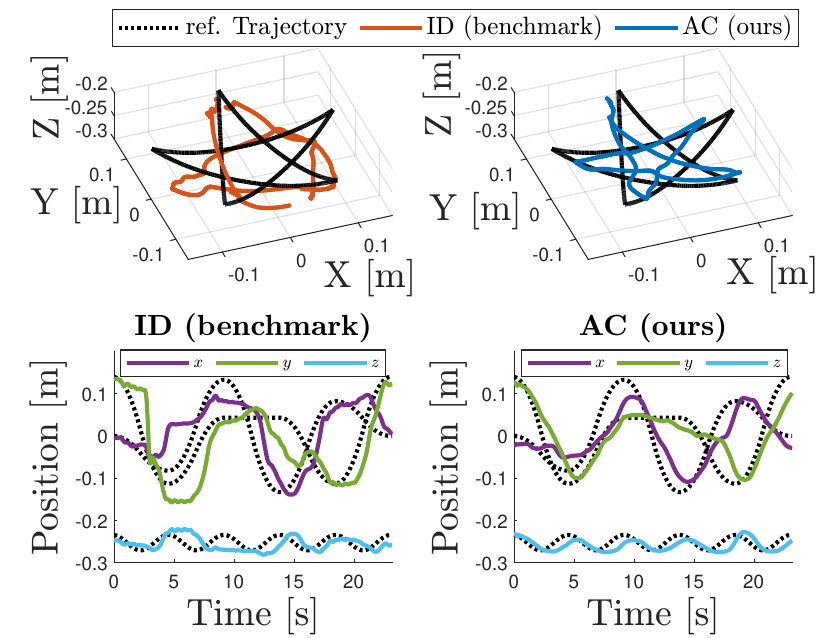}%
\caption{$\SI{25}{\gram}$ payload}%
\end{subfigure}%
\caption{The experimental tracking results for a star-shaped reference trajectory (dotted black line) using our adaptive controller (blue line) and the benchmark controller (red line), both of which carried different payloads. In (a), (b), and (c), the maximum reference velocity and acceleration were set to $v_{max} = \SI{0.05}{\meter\per\second}$ and $a_{max} = \SI{0.01}{\meter\per\second\squared}$, respectively. In (d), (e), and (f), we use $v_{max} = \SI{0.11}{\meter\per\second}$ and $a_{max} = \SI{0.05}{\meter\per\second\squared}$ in the timing law.\label{fig:star_traj}}
\end{figure*}
\vspace{-6pt}
\section{Conclusion and Future Work}
\label{sec:concl}
Our model-based control strategy enables soft continuum robotic arms to track task-space trajectories in a 3D space while carrying an unknown payload. Many parameters, such as stiffness and damping coefficients, must be identified in soft robotic arm models. Moreover, adding a payload to the robot changes the model's dynamic parameters. Our generalizable adaptive controller can update these parameters online for accurate soft robot control. We hope to use these characteristics in future works to assist soft manipulators in performing dynamically loaded tasks, such as picking and placing objects with unknown loads. 
As our adaptive control scheme ignores uncertainties in the actuator mapping matrix or the presence of actuator faults, a fault-tolerant control approach deserves investigation for future work.
\FloatBarrier

\addtolength{\textheight}{-12cm}  
\bibliographystyle{IEEEtran}
\bibliography{references}
\end{document}